\documentclass[runningheads]{llncs}

\usepackage{eccv}

\usepackage{eccvabbrv}

\usepackage{graphicx}
\usepackage{booktabs}

\usepackage[accsupp]{axessibility}  % Improves PDF readability for those with disabilities.

\usepackage{hyperref}
\usepackage{multirow}
\usepackage{amssymb}                       % \checkmark for ablation table
\usepackage{orcidlink}

\begin{document}

\title{Learning Structurally Consistent Representations for Multi-View Radar Semantic Segmentation} 

\titlerunning{HyperRadar}

\author{Ali Zia$^{\dagger{*}}$\inst{1,2}\orcidlink{0000-0003-4819-6666} \and
 Muhammad Umer Ramzan$^{*}$\inst{3}\orcidlink{0009-0002-2760-9724} \and
Abdelwahed Khamis\inst{4}\orcidlink{0000-0002-3475-3479}
\and
Usman Ali\inst{3}\orcidlink{0009-0003-0821-1849}\and
Abdul Rehman\inst{3}\orcidlink{0009-0001-0003-3932}}

\authorrunning{A.~Zia et al.}

\institute{School of Computing, Engineering
and Mathematical Sciences, La Trobe University, Melbourne, Australia 
\and La Trobe Institute for Sustainable Agriculture \& Food (LISAF), Australia 
\and
School of Engineering and Applied Sciences, GIFT University, Pakistan
% \institute{La Trobe University, Melbourne, Australia \and
% GIFT University, Gujranwala, Punjab, Pakistan
% \url{http://www.springer.com/gp/computer-science/lncs} 
\and
Commonwealth Scientific and Industrial Research Organisation (CSIRO), Australia
% CSIRO Data61, Australia\\
\email{A.Zia@latrobe.edu.au, \{umer.ramzan,usmanali,abdulrehman.naseer\}@gift.edu.pk, abdelwahed.khamis@data61.csiro.au}}

\maketitle

\begingroup
\renewcommand\thefootnote{}
\footnotetext{* Equal Contribution. \textdagger{} Corresponding Author.}
\endgroup

\begin{abstract}
Radar sensors provide reliable perception under adverse weather and lighting conditions, but their sparse, noisy, and weakly semantic measurements make dense semantic segmentation challenging. Most existing radar segmentation methods rely on grid-based encodings and pairwise interactions, which struggle to capture the higher-order relational structure formed by multiple radar returns from the same physical object.
We introduce a unified higher-order structural alignment framework for multi-view radar segmentation. The proposed method refines radar feature representations using learnable hypergraphs to capture higher-order dependencies among spatially related responses. To ensure consistency across heterogeneous radar projections, we further align view-specific features using Unbalanced Optimal Transport (UOT), enabling correspondence-free alignment under varying measurement densities and partial observations. An adaptive attention mechanism then fuses complementary radar views while emphasising structurally informative responses under sparsity and noise.
The resulting architecture learns structurally consistent representations across Range Angle (RA), Range Doppler (RD), and Angle Doppler (AD) views and is trained using supervised segmentation together with cross-view consistency regularisation. Experiments on the CARRADA and RADIal benchmarks demonstrate consistent improvements over strong radar-specific baselines, achieving \textbf{63.8\% mIoU on CARRADA and 83.4\% mIoU on RADIal}, improving the previous best methods by \textbf{+1.7 and +2.3 mIoU}, respectively. These results highlight the importance of higher-order relational modelling for robust radar perception.

\end{abstract}

\section{Introduction}
\label{sec:intro}

Reliable scene understanding is a fundamental requirement for autonomous driving, particularly under conditions where perception quality directly affects safety-critical decisions. Among commonly used sensing modalities, radar is especially attractive because it remains operational under adverse weather, low illumination, glare, fog, and other degradations that can severely affect cameras and LiDAR \cite{yao2023radar}. This robustness makes radar an important component of modern perception stacks. However, despite its favourable sensing characteristics, semantic segmentation from radar remains substantially more difficult than in image or point-cloud domains. 
Radar measurements are inherently sparse, noisy, low-resolution, and weakly semantic, which limits the discriminative content available for dense scene parsing \cite{dalbah2024transradar, li2025get}. 
In practice, radar evidence rarely appears as isolated responses; instead, multiple radar returns from the same object form groups of spatially and spectrally related measurements. Conventional convolutional or pairwise models struggle to capture this structure.
For this reason, learning representations that are both structurally reliable and semantically expressive remains a central challenge in radar perception.

A large body of prior work on radar segmentation relies on grid-based encodings or convolutional architectures operating over range-angle, range-Doppler, or related radar tensors \cite{lombacher2017semantic, sless2019road}. While these formulations enable the use of mature dense prediction pipelines, they often inherit an implicit assumption of locally smooth and densely informative observations. This assumption is poorly matched to radar sensing, where returns are irregular, cluttered, and often fragmented across multiple cells. Moreover, radar reflections originating from a single physical object may be distributed across several neighbouring bins due to object extent, multipath effects, motion, and sensor viewpoint. Under such conditions, local convolutions and pairwise interactions are often insufficient to capture the higher-order dependencies needed for reliable semantic reasoning. Consequently, models may overfit to local artefacts, struggle under severe sparsity, and produce unstable segmentations in challenging scenes.

Recent efforts have sought to improve radar understanding through graph-based learning, self-supervision, and contrastive objectives, aiming to better exploit relational structure and reduce reliance on dense annotations \cite{fent2023radargnn, hao2024bootstrapping, xiong2022contrastive}. 
Despite recent advances in radar perception, two limitations remain. First, most methods rely on convolutional or pairwise interactions that cannot explicitly represent groups of radar responses originating from the same object. Second, multi-view radar observations often exhibit unequal measurement densities and partial overlap, making conventional alignment strategies brittle. Addressing both challenges requires representations that capture higher-order structure while remaining robust to incomplete cross-view correspondence.

To address these limitations, we propose a structurally consistent representation learning framework for radar perception that combines hypergraph reasoning with unbalanced optimal transport alignment. Radar returns from the same physical object often appear as groups of spatially and spectrally related responses distributed across multiple cells, a structure that conventional convolutional or pairwise models fail to capture.

We therefore represent radar features using \emph{hypergraphs}, where a single hyperedge connects multiple related measurements or embeddings. This formulation enables higher-order relational modelling and promotes region-level coherence, allowing the network to capture object-level structure and remain robust to sparsity, clutter, and ambiguous radar observations.

While hypergraph reasoning captures intra-view structure, robust segmentation also requires consistent representations across radar views. In practice, different projections often exhibit varying measurement densities and partial observations due to occlusions and view-dependent sparsity. To handle these inconsistencies, we employ \emph{Unbalanced Optimal Transport} (UOT)~\cite{sejourne2023unbalanced} as a correspondence-free alignment mechanism for multi-view radar features. Unlike balanced transport, UOT allows mass variation across distributions, making it well suited for radar data where the number and strength of returns differ between views. Applied to structurally refined features, UOT encourages consistent cross-view organisation while remaining tolerant to missing or spurious measurements.
To further improve feature aggregation under sparse sensing conditions, we incorporate an adaptive attention mechanism that dynamically emphasises informative radar responses while suppressing noise.
Building on these ideas, we introduce \textbf{HyperRadar}, an end-to-end framework for multi-view radar semantic segmentation that integrates higher-order hypergraph reasoning, correspondence-free UOT alignment, and adaptive attention-based refinement. By jointly enforcing intra-view structural coherence and inter-view distributional alignment, HyperRadar produces reliable semantic predictions under sparse and noisy radar observations.
This paper makes the following contributions:
\begin{itemize}
    \item We introduce a \textbf{structurally grounded radar representation framework} that models groups of radar returns using \textbf{hypergraphs}. This formulation enables explicit higher-order relational reasoning and captures object-level structure that cannot be represented through local convolutions or pairwise interactions.

    \item We propose a \textbf{correspondence-free multi-view alignment mechanism based on Unbalanced Optimal Transport (UOT)}, which aligns radar feature distributions across projections with varying measurement densities and partial observations, improving robustness to realistic sensing inconsistencies.

    \item We develop a \textbf{hypergraph-aware feature refinement strategy with adaptive attention}, which enforces coherent region-level representations and selectively emphasises informative radar responses for improved semantic segmentation under sparse and noisy sensing conditions.
\end{itemize}
% This paper makes the following contributions:
% \begin{itemize}
%     \item We introduce a \textbf{structurally grounded radar segmentation framework} that models radar measurements with \textbf{hypergraphs}, enabling explicit higher-order relational reasoning over groups of radar returns rather than relying only on local convolutions or pairwise interactions.

%     \item We propose a \textbf{correspondence-free multi-view alignment module based on Unbalanced Optimal Transport}, which aligns radar feature distributions under varying measurement densities, partial observations, and clutter, making the learned representation more robust to realistic sensing inconsistencies.

%     \item We develop a \textbf{hypergraph-aware consistency learning strategy}, coupled with \textbf{adaptive attention-based feature refinement}, to enforce coherent region-level representations and selectively emphasise structurally informative radar responses for improved semantic segmentation.

%     \item We integrate these components into \textbf{HyperRadar}, an efficient end-to-end architecture for multi-view radar semantic segmentation, and demonstrate \textbf{consistent state-of-the-art performance} on CARRADA and RADIal with a more parameter-efficient design.
% \end{itemize}

Extensive experiments on the CARRADA~\cite{ouaknine2021carrada} and RADIal~\cite{rebut2022raw} benchmarks demonstrate that HyperRadar consistently outperforms strong radar-specific baselines, achieving \textbf{63.8\% mIoU on CARRADA} and \textbf{83.4\% mIoU on RADIal}. This corresponds to improvements of \textbf{+1.7} and \textbf{+2.3 mIoU} over the previous best methods, respectively, highlighting the effectiveness of higher-order relational modelling for radar perception.

\section{Related Work}

Low-cost frequency-modulated continuous-wave (FMCW) radars have been increasingly adopted in machine learning and pattern recognition tasks, including human activity recognition and hand gesture analysis \cite{zhang2020object,zhang2019u,zhang2018latern}. In autonomous driving, however, LiDAR has historically been preferred due to its high-resolution geometric measurements, and several studies have therefore explored radar-LiDAR fusion within point-cloud-oriented pipelines \cite{bansal2022radsegnet,feng2020deep}. Despite these efforts, radar measurements differ fundamentally from LiDAR returns in sensing physics and signal formation. Their low spatial resolution, high sparsity, speckle-like noise, and clutter significantly limit the effectiveness of point-cloud abstractions and architectures designed for denser geometric observations \cite{palffy2020cnn,schumann2021radarscenes}.

To address these limitations, recent work has shifted toward image-like radar representations derived from signal processing in the Range, Azimuth, and Doppler domains. While some datasets provide radar data in point-cloud format \cite{barnes2020oxford,schumann2021radarscenes}, most contemporary automotive radar benchmarks provide either single-view projections such as Range-Azimuth (RA) or Range-Doppler (RD) maps \cite{rebut2022raw,wang2021rodnet}, raw radar signals \cite{rebut2022raw}, or full Range-Azimuth-Doppler (RAD) tensors \cite{ouaknine2021multi,zhang2021raddet}. RAD tensors preserve richer spatial and kinematic structure, but their three-dimensional nature substantially increases memory and computational cost, especially for temporal modelling. Consequently, several methods operate on sliced or projected RAD representations to balance representational richness and efficiency \cite{gao2020ramp,ouaknine2021multi}.

With the emergence of annotated automotive radar datasets such as CARRADA and RADIal \cite{ouaknine2021carrada,rebut2022raw}, a growing number of methods have been proposed for radar semantic segmentation and object detection. Early baselines adapt conventional dense-prediction architectures, including UNet \cite{ronneberger2015u} and DeepLabv3+ \cite{chen2018encoder}. Although effective for natural images, these models are not explicitly designed for the sparsity, noise, and clutter characteristic of radar imagery. TMVANet \cite{ouaknine2021multi} addresses this limitation using a multi-view convolutional architecture with separate encoding, latent interaction, and decoding stages, yielding strong results on both RA and RD segmentation. RAMP-CNN \cite{gao2020ramp}, originally introduced for 3D RAD processing, has also influenced radar perception pipelines, though its reliance on volumetric convolutions limits scalability.

More recent methods have explored attention-based and feature-selective mechanisms to better capture long-range dependencies and salient radar responses. T-RODNet \cite{jiang2022t} employs Swin Transformers \cite{liu2021swin} for RA-based object detection, demonstrating the value of non-local feature aggregation, but it does not jointly model multi-view segmentation. PeakConv \cite{zhang2023peakconv} introduces peak-aware convolution to focus on strong radar reflections and improves over TMVANet, albeit with higher parameter count and computational cost. Related sparse attention mechanisms, such as ReLA \cite{zhang2021sparse}, neighbourhood point attention \cite{xue2022efficient}, and SCAN \cite{xu2022sparse}, further show that selective feature activation can improve efficiency. However, these methods either target point-cloud data or operate primarily through local/pairwise interactions, and thus do not explicitly model the higher-order relational structure induced by multiple radar returns from the same physical object.

Hypergraphs provide a principled generalisation of graphs in which a single hyperedge can connect more than two vertices, making them well-suited to modelling group-wise dependencies and higher-order structure. Early work on spectral learning with hypergraphs showed their usefulness for clustering, classification, and embedding in settings where pairwise graphs are insufficient \cite{zhou2006learning}. More recently, hypergraph neural models such as HGNN \cite{feng2019hypergraph} and HyperGCN \cite{yadati2019hypergcn} have demonstrated that hypergraph-based message passing can effectively capture complex multi-way relations while remaining compatible with deep representation learning. These ideas are particularly relevant for radar perception, where meaningful evidence is often distributed across sets of neighbouring bins or returns associated with the same object, extended surface, or motion pattern. Nevertheless, prior radar segmentation methods predominantly rely on convolutional operators, standard attention, or pairwise interactions, and do not explicitly exploit hypergraph structure for semantic segmentation. Our work builds on this line of research by introducing hypergraph-based reasoning directly into multi-view radar representation learning, enabling region-level structural consistency beyond local or pairwise aggregation.

Optimal Transport (OT) has emerged as a powerful tool for comparing and aligning distributions in a geometrically meaningful manner, with broad impact across machine learning and computer vision \cite{peyre2019computational}. Entropically regularised OT and Sinkhorn-based solvers \cite{cuturi2013sinkhorn} have made OT practical at modern learning scales, while applications such as domain adaptation have shown its effectiveness for correspondence-free alignment between heterogeneous feature distributions \cite{courty2016optimal}. However, classical OT assumes balanced mass preservation, which can be restrictive when observations are incomplete, corrupted, or unevenly distributed. Unbalanced Optimal Transport (UOT) relaxes this assumption by allowing discrepancies in total mass, thereby improving robustness to outliers, missing observations, and partial overlap between distributions \cite{sejourne2023unbalanced}. This property is especially desirable for multi-view radar, where different views often contain unequal numbers of valid returns due to occlusion, sparsity, clutter, and viewpoint-dependent visibility. In contrast, our method leverages UOT to align multi-view radar feature distributions without requiring explicit correspondences, \textcolor{black}{a design choice we detail and validate in Sec.~\ref{sec:uot}.}

In summary, prior radar segmentation methods have improved feature extraction through multi-view convolutions, transformers, and sparse attention, but they remain limited in their ability to model higher-order radar structure and to align heterogeneous multi-view observations under unequal measurement densities. Our approach addresses these two gaps jointly by combining hypergraph-based structural reasoning with unbalanced optimal transport, yielding a representation that is better matched to the physical and statistical characteristics of automotive radar data.

% \begin{figure*}[htbp]
%         \centering
%         \includegraphics[width=\textwidth]{ECCV.pdf}
%         \caption{The proposed framework processes three radar modalities, Range–Doppler (RD), Range–Angle (RA), and Angle–Doppler (AD), using the encoders to learn domain-aware features. These features are structurally aligned across modalities using unbalanced optimal transport to capture geometric consistency between heterogeneous radar views. \textcolor{red}{The Adaptive Directional Attention (ADA) Module is adopted from the TransRadar paper.} Overall, the architecture effectively fuses complementary radar information while preserving structural coherence across modalities.}
%         \label{fig:architecture}
%     \end{figure*}

\begin{figure*}[htbp]
        \centering
        \includegraphics[width=\textwidth]{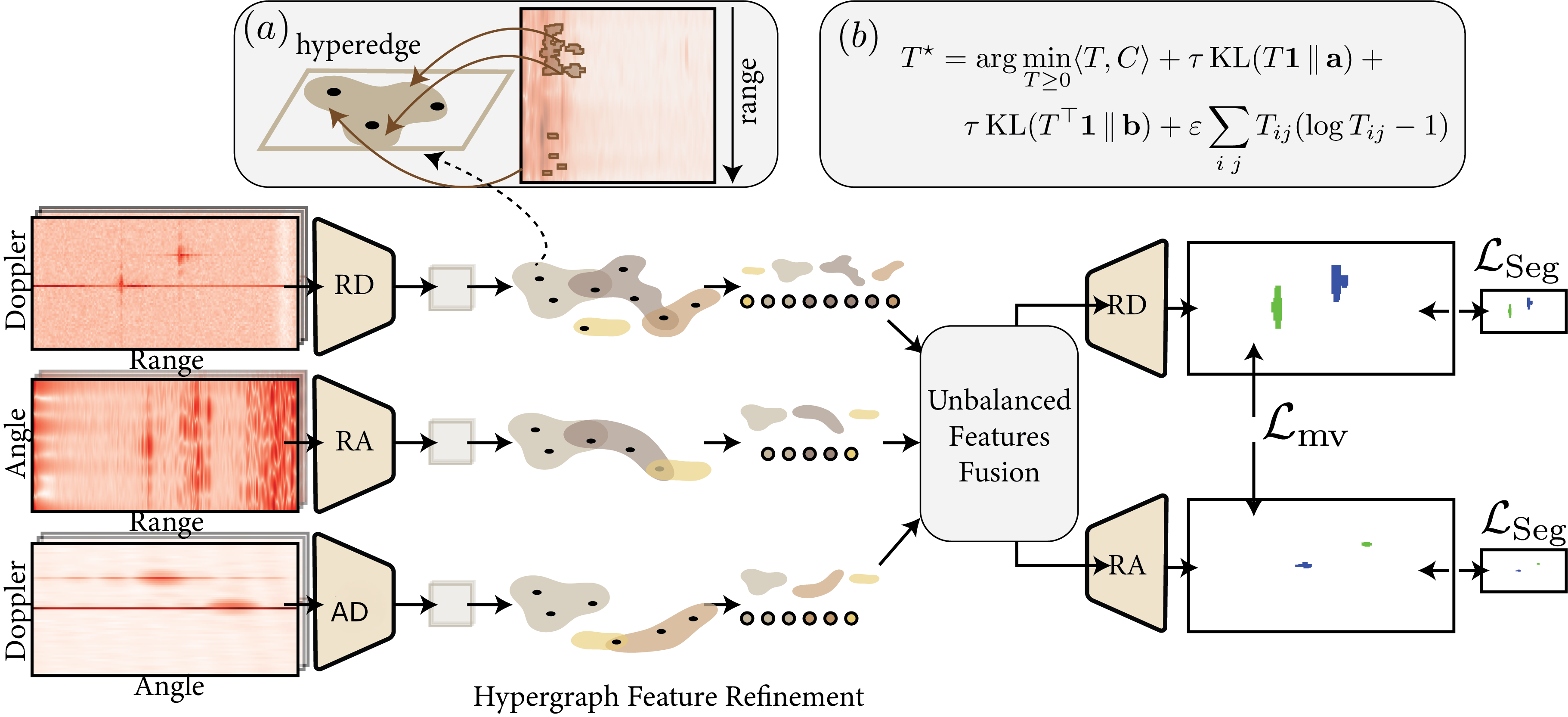}
        \caption{Architecture of the proposed HyperRadar framework. Three radar views (RD, RA, AD) are processed by independent encoders and refined using view-specific learnable hypergraphs. (a) Hypergraph refinement: for an extended object like a car with distant reflections spanning multiple range cells, the hypergraph captures these non-local dependencies and groups them into a single hyperedge. (b) Unbalanced Optimal Transport (UOT) aligns the refined features across views. The fused tokens are refined via adaptive attention and subsequently decoded into dense segmentation masks. The network is trained end-to-end using joint segmentation and cross-view consistency objectives.
        % The fused tokens are refined by the adaptive attention and decoded into dense masks, trained end-to-end with joint segmentation and cross-view consistency objectives.
        }
        \label{fig:architecture}
    \end{figure*}

\section{Methodology}
\label{sec:method}

\subsection{Overview}
Radar semantic segmentation is challenging due to sparse, noisy, and weakly semantic measurements, as well as view-dependent missing returns. While local convolutions and pairwise fusion capture short-range patterns, they are limited in modelling the higher-order structure induced by multiple returns from the same physical object. We propose \textbf{HyperRadar}, a compact multi-view framework that (i) extracts view-specific features from three RAD projections, (ii) performs \emph{per-view} learnable hypergraph refinement to capture higher-order dependencies, (iii) aligns and fuses refined features across views via \emph{unbalanced optimal transport} (UOT), and (iv) decodes RD and RA semantic masks with cross-view consistency regularisation. The overall pipeline is illustrated in Figure~\ref{fig:architecture}.

Given a radar cube at time $t_0$, we form three 2D projections from the 3D Range--Angle--Doppler (RAD) representation: Range--Doppler (RD), Range--Angle (RA), and Angle--Doppler (AD). For each view $v \in \{\mathrm{RD}, \mathrm{RA}, \mathrm{AD}\}$, the input is $x^{(v)} \in \mathbb{R}^{T \times H_v \times W_v}$, where $T$ is the temporal window and $H_v, W_v$ are the spatial dimensions. A view-specific encoder maps each input to a latent representation:
\begin{equation}
F^{(v)} = E_v(x^{(v)}) \in \mathbb{R}^{C \times H_d^{(v)} \times W_d^{(v)}},
\label{eq:encoder}
\end{equation}
where $C$ is the latent channel dimension. Flattening $F^{(v)}$ yields tokens $X^{(v)} \in \mathbb{R}^{N_v \times C}$ with $N_v = H_d^{(v)} W_d^{(v)}$, where each row corresponds to one latent location.

\subsection{Learnable Hypergraph-Based Feature Refinement}
\label{sec:hypergraph}
For each view $v$, we refine latent tokens using a learnable hypergraph to capture higher-order interactions among spatially and spectrally related radar responses. We define a hypergraph $\mathcal{G}^{(v)}=(\mathcal{V},\mathcal{E}^{(v)})$ over the $N_v$ tokens, where $\mathcal{V}=\{1,\dots,N_v\}$ and $\mathcal{E}^{(v)}=\{e_1,\dots,e_M\}$ contains $M$ learnable hyperedges. Instead of fixed neighbourhoods, node-to-hyperedge associations are learned via a soft incidence matrix
\[
H^{(v)}=\mathrm{softmax}(X^{(v)}W_h)\in\mathbb{R}^{N_v\times M},
\]
where $W_h\in\mathbb{R}^{C\times M}$ is trainable and the softmax is applied along the hyperedge dimension so each token distributes its membership across hyperedges. Hyperedge embeddings are computed by weighted aggregation $Z^{(v)}=(H^{(v)})^\top X^{(v)}$, transformed by a lightweight two-layer mapping $\phi(\cdot)$, and projected back to tokens through the same incidence structure. The refinement is expressed as
\begin{equation}
X_{\mathrm{hg}}^{(v)} = X^{(v)} + \gamma\, H^{(v)} \phi\!\left((H^{(v)})^\top X^{(v)}\right),
\label{eq:hypergraph}
\end{equation}
where $\gamma$ is a learnable scalar. This residual hypergraph update allows tokens participating in the same higher-order structures to exchange contextual information, improving robustness to clutter, sparsity, and fragmented returns.

\subsection{Unbalanced Optimal Transport for Multi-View Feature Fusion}
\label{sec:uot}
Although hypergraph refinement improves within-view structural coherence, features remain misaligned across views due to unequal measurement densities and partial observations. We therefore align and fuse hypergraph-refined features across views using unbalanced optimal transport (UOT), which relaxes strict mass preservation and is more suitable for radar sparsity.

For a pair of views $(p,q)$, let $X_{\mathrm{hg}}^{(p)}\in\mathbb{R}^{N_p\times C}$ and $X_{\mathrm{hg}}^{(q)}\in\mathbb{R}^{N_q\times C}$ denote refined tokens. We define the transport cost matrix $C^{(p,q)}\in\mathbb{R}^{N_p\times N_q}$ by
\[
C^{(p,q)}_{ij}=\|x^{(p)}_i-x^{(q)}_j\|_2^2,
\]
where $x^{(p)}_i$ and $x^{(q)}_j$ are token embeddings from views $p$ and $q$. With reference masses $\mathbf{a}=\frac{1}{N_p}\mathbf{1}\in\mathbb{R}^{N_p}$ and $\mathbf{b}=\frac{1}{N_q}\mathbf{1}\in\mathbb{R}^{N_q}$, we solve the entropically regularised UOT problem
\begin{equation}
T^{(p,q)} = \arg\min_{T \ge 0}
\langle T, C^{(p,q)} \rangle
+ \tau\, \mathrm{KL}(T\mathbf{1}\,\|\,\mathbf{a})
+ \tau\, \mathrm{KL}(T^\top\mathbf{1}\,\|\,\mathbf{b})
+ \varepsilon \sum_{i,j} T_{ij}(\log T_{ij}-1),
\label{eq:uot}
\end{equation}
where $\tau$ controls marginal relaxation, $\varepsilon$ is the entropic regularisation weight, and $\mathrm{KL}(\cdot\|\cdot)$ denotes Kullback--Leibler divergence. We compute $T^{(p,q)}$ using a differentiable generalised Sinkhorn solver.
\textcolor{black}{The mass imbalance is inherent to radar physics: For a radar cube $\mathcal{S}[r,d,l]$ with range $r$, Doppler $d$, and antenna element $l$, the RD view integrates $\sum_l|\mathcal{S}[r,d,l]|^2$ over the full antenna array while RA integrates over a narrower aperture, causing a car's ${\sim}3$-bin RD peak to smear into $15{+}$ RA bins and violating balanced-OT mass conservation. UOT's marginal penalty $\tau$ absorbs this mismatch, we use 30 Sinkhorn iterations on normalised downsampled latent tokens, with hyperparameter sensitivity reported in Sec.~\ref{quatitative}.}

We obtain an aligned contribution from view $q$ to view $p$ by transporting features:
\begin{equation}
\widehat{X}^{(p\leftarrow q)} = T^{(p,q)} X_{\mathrm{hg}}^{(q)} \in \mathbb{R}^{N_p\times C}.
\label{eq:transport}
\end{equation}
The \emph{unbalanced features fusion} module then aggregates aligned evidence with the native representation:
\begin{equation}
X_{\mathrm{fused}}^{(p)} = X_{\mathrm{hg}}^{(p)} + \sum_{q\in\mathcal{N}(p)} \beta_{p,q}\,\widehat{X}^{(p\leftarrow q)},
\label{eq:fusion_uot}
\end{equation}
where $\mathcal{N}(p)$ denotes the other views and $\beta_{p,q}$ are learnable scalar fusion weights. In our setting, we form fused representations for the supervised output views $p\in\{\mathrm{RD},\mathrm{RA}\}$ using contributions from the remaining views.

\subsection{Adaptive Attention and Decoding}
\label{sec:attention_dec}
The fused representations remain subject to varying reliability across scenes and views. We therefore apply an adaptive attention mechanism before decoding to emphasise informative responses. In practice, this module can be applied multiple times in sequence to progressively refine the fused features. For each supervised view $p\in\{\mathrm{RD},\mathrm{RA}\}$, we compute a global descriptor by average pooling $g^{(p)}=\frac{1}{N_p}\sum_{i=1}^{N_p}X^{(p)}_{\mathrm{fused},i}\in\mathbb{R}^{C}$ and obtain channel-wise gates via a small function $\psi(\cdot)$. The attended features are
\begin{equation}
\widetilde{X}^{(p)} = \psi(g^{(p)}) \odot X_{\mathrm{fused}}^{(p)},
\label{eq:attn}
\end{equation}
where $\odot$ denotes channel-wise multiplication with broadcasting over tokens. The attended tokens are reshaped back to a latent map and decoded by shallow view-specific decoders:
\begin{equation}
S^{(p)} = D_p\!\left(\mathrm{reshape}(\widetilde{X}^{(p)})\right),
\label{eq:decoder}
\end{equation}
where $S^{(p)}$ are dense semantic logits for view $p$.

\subsection{Training Objectives}
\label{sec:loss}
We train the network end-to-end using supervised segmentation together with cross-view consistency between RD and RA. For each supervised view $p\in\mathcal{V}_{\mathrm{out}}=\{\mathrm{RD},\mathrm{RA}\}$, the segmentation loss combines cross-entropy and soft Dice:
\[
\mathcal{L}_{\mathrm{seg}}^{(p)}=\mathcal{L}_{\mathrm{ce}}^{(p)}+\lambda_{\mathrm{dice}}\mathcal{L}_{\mathrm{dice}}^{(p)},
\qquad
\mathcal{L}_{\mathrm{seg}}=\sum_{p\in\mathcal{V}_{\mathrm{out}}}\mathcal{L}_{\mathrm{seg}}^{(p)}.
\]
Here, $\mathcal{L}_{\mathrm{dice}}$ is computed on softmax-normalised class probabilities to mitigate foreground--background imbalance.

To enforce agreement between RD and RA predictions along their shared range dimension, let $P^{(p)}=\mathrm{softmax}(S^{(p)})$ denote class probabilities. We extract range-wise confidence profiles using a max operator over the non-shared dimension:
\[
Q^{(\mathrm{RA})}=\max_{\text{azimuth}} P^{(\mathrm{RA})}, \qquad
Q^{(\mathrm{RD})}=\mathrm{Rot}_{180}\!\left(\max_{\text{doppler}} P^{(\mathrm{RD})}\right),
\]
where $\mathrm{Rot}_{180}(\cdot)$ aligns the RD range axis with the RA convention. We define two complementary regularisers:
\begin{equation}
\mathcal{L}_{\mathrm{coh}} = \|Q^{(\mathrm{RD})} - Q^{(\mathrm{RA})}\|_2^2,
\qquad
\mathcal{L}_{\mathrm{mv}} = \mathrm{Huber}\!\left(Q^{(\mathrm{RD})},\,Q^{(\mathrm{RA})}\right).
\label{eq:consistency}
\end{equation}
The final objective is
\begin{equation}
\mathcal{L} = \mathcal{L}_{\mathrm{seg}} + \lambda_{\mathrm{coh}}\mathcal{L}_{\mathrm{coh}} + \lambda_{\mathrm{mv}}\mathcal{L}_{\mathrm{mv}},
\label{eq:final_loss}
\end{equation}
where $\lambda_{\mathrm{dice}}$, $\lambda_{\mathrm{coh}}$, and $\lambda_{\mathrm{mv}}$ control the contribution of each term. All modules are optimised jointly in a single end-to-end training procedure.

\section{Experiments}

\subsection{Datasets}

We evaluate the proposed method on two public automotive radar benchmarks: CARRADA~\cite{ouaknine2021carrada} and RADIal~\cite{rebut2022raw}. CARRADA is used as the primary benchmark for multi-class radar semantic segmentation, while RADIal is used to assess generalisation under a different sensing setup and to compare with prior work on both segmentation and detection.

\textbf{CARRADA:}
CARRADA~\cite{ouaknine2021carrada} contains synchronised radar--camera recordings collected in urban driving scenes, with 12,666 annotated frames. Labels are generated semi-automatically and provided directly in the radar domain for the Range--Angle (RA) and Range--Doppler (RD) views. The dataset includes four classes: \emph{pedestrian}, \emph{cyclist}, \emph{car}, and \emph{background}. Radar observations are represented using 2D projections (RA, RD, and Angle--Doppler (AD)) derived from the original 3D RAD tensor. The RA maps are of size $1 \times 256 \times 256$, while the RD and AD maps are $1 \times 256 \times 64$. Following prior work, we use these 2D projections to reduce computational cost while preserving complementary spatial and kinematic information.

\textbf{RADIal:}
RADIal~\cite{rebut2022raw} is a higher-resolution automotive radar benchmark with 8,252 annotated frames. In contrast to CARRADA, it provides a single radar representation and does not support multi-view radar inputs. Its annotations are obtained by projecting labels from aligned RGB images into the radar domain rather than being directly defined in radar coordinates. The radar input has resolution $32 \times 512 \times 256$, reflecting the dataset’s finer-grained spatial representation. RADIal considers two semantic categories, \emph{free space} and \emph{occupied space} (vehicles), and is commonly used for both segmentation and detection evaluation.

\subsection{Evaluation Metrics}

We follow standard evaluation protocols used in radar perception. For semantic segmentation, we report Intersection over Union (IoU) and Dice (F1) scores, together with their class-wise means. Mean IoU (mIoU) is the primary metric on RADIal. Reporting both mIoU and Dice provides a balanced assessment of region overlap and class-level prediction quality. For object detection on RADIal, we follow prior work~\cite{rebut2022raw} and report Average Precision (AP), Average Recall (AR), and box regression errors.

\subsection{Implementation Details}

All models are implemented in \texttt{PyTorch} and trained on a single NVIDIA RTX 5090 GPU. On CARRADA, we use a batch size of 6 and input sequences of 5 consecutive radar frames. Optimisation is performed with Adam~\cite{kingma2014adam} using an initial learning rate of $1\times10^{-4}$. We employ an exponential learning rate schedule with a decay step of 10 epochs. The final configuration uses $8\times$ cascaded blocks of the proposed adaptive attention module. During inference, the batch size is set to 1 while keeping the same temporal context.

For RADIal, we replace the original FFTRadNet~\cite{rebut2022raw} backbone with the proposed architecture while retaining its segmentation and detection heads. \textcolor{black}{Since RADIal is single-view, the cross-view UOT module, AD encoder, and consistency losses ($\mathcal{L}_{\mathrm{coh}}$, $\mathcal{L}_{\mathrm{mv}}$) are disabled, only the single-view encoder, hypergraph refinement, adaptive attention, and FFTRadNet heads are retained, isolating backbone generalisability from the multi-view design.} Unless otherwise stated, we use the same optimisation strategy as on CARRADA to ensure a fair comparison across datasets.

%%%%%%%%%%%%%%%%%%%%%%%%%%%%%%%%%%%%%%%%%%%%%%%%%%%%%%%%%%%%%%%%%%%%%%%%%%%%%
\begin{table}[t]
\centering
\caption{Semantic segmentation performance on the test split of the CARRADA dataset, shown for the RD (Range-Doppler) and RA (Range-Angle) views. Columns from left to right are the view (RD/RA), the name of the method, the intersection-over-union (IoU) score of the four different classes with their mean, and the Dice score for the same classes. Best results are shown in \textcolor{black}{\textbf{black bold}}, and second-best results are highlighted in \textcolor{blue}{\textbf{bold blue}}. \textcolor{black}{The Background class is near-saturated across most methods, all values tied at the column maximum are jointly highlighted.}}
\label{tab:performance}
\resizebox{\textwidth}{!}{
\begin{tabular}{l | c | c c c c c | c c c c c}
\toprule
\textbf{View} & \textbf{Method} & \multicolumn{5}{c|}{\textbf{IoU (\%)}} & \multicolumn{5}{c}{\textbf{Dice (\%)}}\\
\cline{3-12}
& & Bkg. & Ped. & Cycl. & Car & mIoU & Bkg. & Ped. & Cycl. & Car & mDice \\
\midrule
\multirow{9}{*}{\textbf{RD}} 
& FCN-8s \cite{long2015fully} & \textcolor{black}{\textbf{99.7}} & 47.7 & 18.7 & 52.9 & 54.7 & 99.8 & 24.8 & 16.5 & 26.9 & 66.3 \\
& U-Net \cite{ronneberger2015u} & \textcolor{black}{\textbf{99.7}} & 51.1 & \textbf{33.4} & 37.7 & 55.4 & 99.8 & 67.5 & \textbf{50.0} & 54.7 & 68.0 \\
& DeepLabv3+ \cite{chen2018encoder} & \textcolor{black}{\textbf{99.7}} & 43.2 & 11.2 & 49.2 & 50.8 & \textbf{99.9} & 60.3 & 20.2 & 66.0 & 61.6 \\
& RSS-Net \cite{kaul2020rss} & 99.3 & 0.1 & 4.1 & 25.0 & 32.1 & 99.7 & 0.2 & 7.9 & 40.0 & 36.9 \\
& RAMP-CNN \cite{gao2020ramp} & \textcolor{black}{\textbf{99.7}} & 48.8 & 23.2 & 54.7 & 56.6 & \textcolor{black}{\textbf{99.9}} & 65.6 & 37.7 & 70.8 & 68.5 \\
& MVNet \cite{ouaknine2021multi} & 98.0 & 0.0 & 3.8 & 14.1 & 29.0 & 99.0 & 0.0 & 7.3 & 24.8 & 32.8 \\
& TMVA-Net \cite{ouaknine2021multi} & \textcolor{black}{\textbf{99.7}} & 52.6 & 29.0 & 53.4 & 58.7 & 99.8 & 68.9 & 45.0 & 69.6 & 70.9 \\
& PeakConv \cite{zhang2023peakconv} & - & - & - & - & 60.7 & - & - & - & - & 72.5 \\
& TransRadar \cite{dalbah2024transradar} & \textcolor{black}{\textbf{99.7}} & 56.7 & 30.2 & 61.7 & 62.1 & 99.8 & 72.4 & 46.4 & \textcolor{blue}{\textbf{76.3}} & 73.4 \\
\cline{2-12}
& \textbf{HyperRadar (OT)} & \textcolor{black}{\textbf{99.7}} & 57.0 & 30.6 & 62.1 & 62.4 & 99.8 & 72.6 & 46.8 & 75.8 & 73.8 \\
& \textbf{HyperRadar (Hypergraph)} & \textcolor{black}{\textbf{99.7}} & \textcolor{blue}{\textbf{57.2}} & 30.9 & \textcolor{blue}{\textbf{62.3}} & \textcolor{blue}{\textbf{62.5}} & 99.8 & \textcolor{blue}{\textbf{72.9}} & 47.1 & 76.1 & \textcolor{blue}{\textbf{74.0}} \\
& \textbf{HyperRadar (HyperGraph + OT)} & \textbf{99.7} & \textbf{59.2} & \textcolor{blue}{\textbf{32.4}} & \textbf{63.8} & \textbf{63.8} & \textcolor{black}{99.8} & \textbf{74.5} & \textcolor{blue}{\textbf{48.9}} & \textbf{77.6} & \textbf{75.2} \\
\midrule
\multirow{9}{*}{\textbf{RA}} 
& FCN-8s \cite{long2015fully} & 99.8 & 14.8 & 0.0 & 23.3 & 34.5 & 99.9 & 25.8 & 0.0 & 37.8 & 40.9 \\
& U-Net \cite{ronneberger2015u} & 99.8 & 22.4 & \textbf{8.8} & 0.0 & 32.8 & 99.9 & 25.8 & 0.0 & 37.8 & 40.9 \\
& DeepLabv3+ \cite{chen2018encoder} & \textcolor{black}{\textbf{99.9}} & 3.4 & 5.9 & 21.8 & 32.7 & 99.9 & 6.5 & 11.1 & 35.7 & 38.3 \\
& RSS-Net \cite{kaul2020rss} & 99.5 & 7.3 & 5.6 & 15.8 & 32.1 & 99.8 & 13.7 & 10.5 & 27.4 & 37.8 \\
& RAMP-CNN \cite{gao2020ramp} & 99.8 & 1.7 & 2.6 & 7.2 & 27.9 & 99.9 & 3.4 & 5.1 & 13.5 & 30.5 \\
& MVNet \cite{ouaknine2021multi} & 98.8 & 0.1 & 1.1 & 6.2 & 26.8 & 99.0 & 0.0 & 7.3 & 24.8 & 28.5 \\
& TMVA-Net \cite{ouaknine2021multi} & 99.8 & 26.0 & 8.6 & 30.7 & 41.3 & 99.9 & 41.3 & \textbf{15.9} & 47.0 & 51.0 \\
& PeakConv \cite{zhang2023peakconv} & - & - & - & - & 42.9 & - & - & - & - & \textcolor{blue}{\textbf{53.3}} \\
& TransRadar \cite{dalbah2024transradar} & \textcolor{black}{\textbf{99.9}} & 29.9 & 6.5 & 35.3 & 42.9 & 99.9 & 46.0 & 12.2 & 52.2 & 52.6 \\
\cline{2-12}
& \textbf{HyperRadar (OT)} & \textcolor{black}{\textbf{99.9}} & 30.1 & 6.8 & 35.6 & 43.1 & 99.9 & 46.3 & 12.6 & 52.5 & 52.8 \\
& \textbf{HyperRadar (Hypergraph)} & \textcolor{black}{\textbf{99.9}} & \textcolor{blue}{\textbf{30.4}} & 7.1 & \textcolor{blue}{\textbf{35.9}} & \textcolor{blue}{\textbf{43.3}} & 99.9 & \textcolor{blue}{\textbf{46.7}} & 12.9 & \textcolor{blue}{\textbf{52.9}} & 53.1 \\
& \textbf{HyperRadar (HyperGraph + OT)} & \textbf{99.9} & \textbf{31.8} & \textcolor{blue}{\textbf{8.4}} & \textbf{37.5} & \textbf{44.4} & \textbf{99.9} & \textbf{48.2} & \textcolor{blue}{\textbf{14.6}} & \textbf{54.4} & \textbf{54.3} \\
\bottomrule
\end{tabular}
}
\end{table}

%%%%%%%%%%%%%%%%%%%%%%%%%%%%%%%%%%%%%%%%%%%%%%%%%%%%%%%%%%%%%%%%%%%%%%%%%%%%%%%%%%
\subsection{Quantitative Results}\label{quatitative}
\subsubsection{Semantic Segmentation Results on CARRADA:}
Table~\ref{tab:performance} reports semantic segmentation results on the CARRADA test split for both RD and RA views. Across both views, the full \textbf{HyperRadar} model achieves the best overall performance, with the clearest gains on sparse foreground classes such as \emph{pedestrian} and \emph{cyclist}, indicating the benefit of combining higher-order structural modelling with cross-view alignment.

In the RD view, HyperRadar achieves the best mIoU/\mbox{mDice} of \textbf{63.8/75.2}\%, improving over TransRadar (62.1/73.4\%) by \textbf{+1.7/+1.8} points. Gains are most evident for foreground classes, with pedestrian, cyclist, and car IoU improving to \textbf{59.2}\%, \textbf{32.4}\%, and \textbf{63.8}\%, respectively. In the more challenging RA view, HyperRadar again yields the best mIoU/\mbox{mDice} of \textbf{44.4/54.3}\%, compared with 42.9/52.6\% for TransRadar. The largest relative gain is for cyclists, where IoU increases from 6.5\% to \textbf{8.4\%}, while pedestrian and car IoU also improve to \textbf{31.8}\% and \textbf{37.5}\%.

The ablation results further show that the two proposed components are complementary. In RD, OT alone and hypergraph refinement alone achieve 62.4\% and 62.5\% mIoU, respectively, while their combination reaches \textbf{63.8\%}. The same trend holds in RA (43.1\%, 43.3\%, and \textbf{44.4\%}), confirming that hypergraph refinement improves intra-view structural coherence, whereas OT strengthens cross-view consistency; together they deliver the most reliable segmentation across both radar views.

\begin{table}[t]
\centering
\caption{Results on the RADIal benchmark. We compare different backbones for joint detection and segmentation. AP and AR denote average precision and average recall, respectively; $R$ and $A$ denote range and angle errors. Best results are shown in \textcolor{black}{\textbf{black bold}}, and second-best results are highlighted in \textcolor{blue}{\textbf{bold blue}}.}
\label{tab:radial_results}
\setlength{\tabcolsep}{4pt}
\renewcommand{\arraystretch}{1.0}
\small
\resizebox{\linewidth}{!}{%
\begin{tabular}{lccccc}
\toprule
\textbf{Backbone} & \textbf{AP (\%) $\uparrow$} & \textbf{AR (\%) $\uparrow$} & \textbf{$R$ (m) $\downarrow$} & \textbf{$A$ ($^\circ$) $\downarrow$} & \textbf{mIoU (\%) $\uparrow$} \\
\midrule
Pixor \cite{yang2018pixor}     & 96.6 & 81.7 & \textbf{0.10} & 0.20 & -- \\
FFTRadNet \cite{rebut2022raw} & 96.8 & 82.2 & \textcolor{blue}{\textbf{0.11}} & 0.17 & 74.0 \\
C-M DNN \cite{jin2023cross}   & 96.9 & 83.5 & -- & -- & 80.4 \\
TransRadar \cite{dalbah2024transradar} & \textcolor{blue}{\textbf{97.3}} & \textcolor{blue}{\textbf{98.4}} & \textcolor{blue}{\textbf{0.11}} & \textcolor{blue}{\textbf{0.10}} & \textcolor{blue}{\textbf{81.1}} \\
\textbf{HyperRadar} & \textbf{98.5} & \textbf{98.9} & \textbf{0.10} & \textbf{0.08} & \textbf{83.4} \\
\bottomrule
\end{tabular}%
}
\end{table}

\subsubsection{Semantic Segmentation Results on RADIal:}

Table~\ref{tab:radial_results} reports results on the RADIal benchmark for both detection and segmentation. The proposed \textbf{HyperRadar} achieves the best overall performance across all reported metrics, demonstrating that the learned representation generalises effectively beyond CARRADA and remains robust under a different sensing configuration.

Compared with the strongest prior method, TransRadar, HyperRadar improves AP from 97.3\% to \textbf{98.5\%} and AR from 98.4\% to \textbf{98.9\%}. At the same time, it reduces the angle error from 0.10$^\circ$ to \textbf{0.08$^\circ$}, while matching the best reported range error of \textbf{0.10\,m}. These gains indicate that the proposed backbone not only improves detection confidence and recall, but also yields more precise localisation.

For segmentation, HyperRadar achieves the highest mIoU of \textbf{83.4\%}, outperforming TransRadar (81.1\%) by \textbf{+2.3} points and C-M DNN (80.4\%) by \textbf{+3.0} points. The margin over FFTRadNet is larger still (\textbf{+9.4} points), highlighting the benefit of the proposed representation over earlier radar-specific backbones. Since RADIal provides a single-view radar setting, these improvements show that the proposed architecture remains effective even when the full multi-view configuration is not available.

Overall, the consistent gains across AP, AR, localisation error, and mIoU show that HyperRadar provides a stronger and more transferable backbone for radar perception. The results suggest that the proposed structural refinement and feature alignment strategy improve both dense semantic understanding and downstream detection quality.

\subsubsection{Computational Cost and Runtime:}
We also compare the computational footprint of \textbf{HyperRadar} against TransRadar in terms of GFLOPs, inference throughput (FPS), and peak GPU memory. HyperRadar requires slightly fewer computations (\textbf{1132.13} vs.\ 1138.57 GFLOPs) and lower GPU memory (\textbf{3.18} vs.\ 3.25 GB), while also achieving higher inference speed (\textbf{10.55} vs.\ 9.6 FPS). These results show that the proposed structural modelling improves segmentation performance without increasing computational cost and, in practice, offers a modest efficiency gain over the strongest baseline.

\textcolor{black}{Table~\ref{tab:ablation}(a) shows each component of HyperRadar is complementary: the attention-only baseline obtains \textbf{52.5} Avg.; adding UOT or HG individually reaches \textbf{52.8} and \textbf{52.9}; the full HG\,+\,UOT\,+\, Attention model achieves \textbf{54.1} Avg.\ (\textbf{63.8/44.4} RD/RA), a \textbf{+1.6} Avg.\ gain. Table~\ref{tab:ablation}(b) confirms robustness: $\varepsilon\in\{0.01,0.05,0.10\}$ and $\tau\in\{0.5,0.8,1.0\}$ keep Avg.\ mIoU within \textbf{53.6--54.1} with no loss divergence, and performance plateaus at $M{=}8$ hyperedges, ruling out hyperedge-count sensitivity.}

% ── Camera-Ready Addition──────
\textcolor{black}{\subsubsection{Component Ablation and UOT Sensitivity.}}
\label{sec:ablation}

\begin{table}[t]
\caption{\textcolor{black}{Component ablation and UOT sensitivity on CARRADA.
  Default: $\varepsilon{=}0.05$, $\tau{=}0.8$; no loss divergence across all settings.}}
\label{tab:ablation}
\centering
\renewcommand{\arraystretch}{0.85}
{\color{black}\scriptsize
\begin{minipage}[t]{0.47\linewidth}
  \centering
  \textbf{(a) Component ablation}\\[1pt]
  \resizebox{\linewidth}{!}{%
    \begin{tabular}{ccc|ccc}
      \toprule
      HG & UOT & Attn. & RD   & RA   & Avg. \\
      \midrule
      --          & --          & \checkmark  & 62.1         & 42.9         & 52.5         \\
      --          & \checkmark  & --          & 61.7         & 42.2         & 52.0         \\
      --          & \checkmark  & \checkmark  & 62.4         & 43.1         & 52.8         \\
      \checkmark  & --          & --          & 61.9         & 42.4         & 52.2         \\
      \checkmark  & --          & \checkmark  & 62.5         & 43.3         & 52.9         \\
      \checkmark  & \checkmark  & --          & 63.0         & 43.7         & 53.4         \\
      \checkmark  & \checkmark  & \checkmark  & \textbf{63.8}& \textbf{44.4}& \textbf{54.1}\\
      \bottomrule
    \end{tabular}%
  }
\end{minipage}%
\hfill
\begin{minipage}[t]{0.47\linewidth}
  \centering
  \textbf{(b) UOT sensitivity}\\[1pt]
  \resizebox{\linewidth}{!}{%
    \begin{tabular}{cc|ccc}
      \toprule
      $\varepsilon$ & $\tau$ & RD   & RA   & Avg. \\
      \midrule
      0.01 & 0.8 & 63.4 & 44.0 & 53.7 \\
      \textbf{0.05} & \textbf{0.8} & \textbf{63.8} & \textbf{44.4} & \textbf{54.1}\\
      0.10 & 0.8 & 63.5 & 44.1 & 53.8 \\
      0.05 & 0.5 & 63.3 & 43.8 & 53.6 \\
      0.05 & 1.0 & 63.6 & 44.2 & 53.9 \\
      \bottomrule
    \end{tabular}%
  }
\end{minipage}
}
\end{table}
%%%%%%%%%%%%%%%%%%%%%%%%%%%%%%%%%%%%%%%

\begin{figure*}[t]
    \centering
    \includegraphics[width=0.99\textwidth]{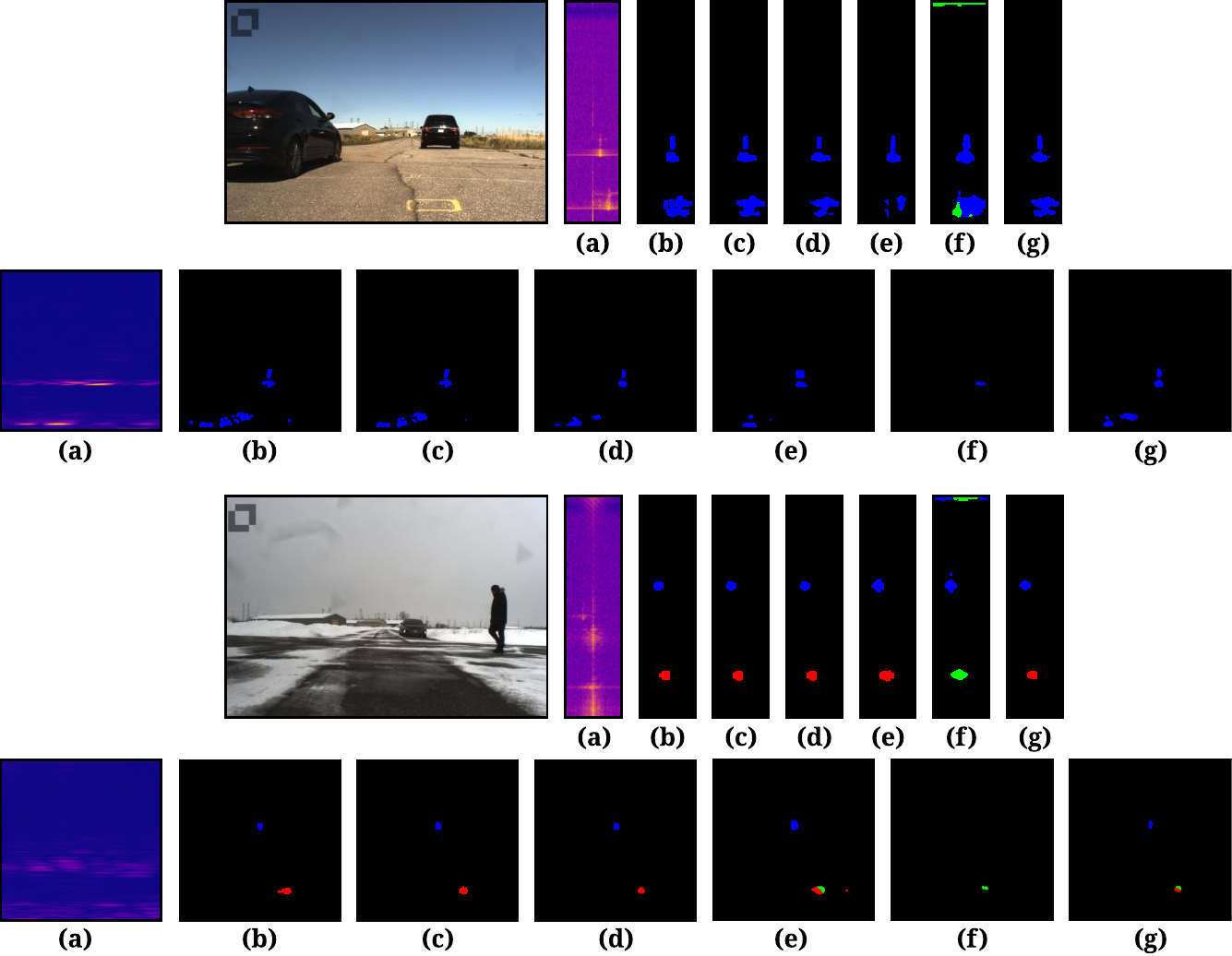}
   \caption{Qualitative comparisons on two test scenes from the CARRADA test split, showing the RGB camera view alongside semantic segmentation outputs from different methods. For each scene, the top row corresponds to the RD view and the bottom row to the RA view. (a) RD/RA inputs, (b) ground truth, (c) Our method, (d) TransRadar \cite{dalbah2024transradar}, (e) TMVA-Net \cite{ouaknine2021multi}, (f) MVNet \cite{ouaknine2021multi}, and (g) U-Net \cite{ronneberger2015u}. All RD outputs are rotated for visual consistency. Colours denote semantic classes: \textcolor{blue}{blue for Car}, \textcolor{green}{green for Cyclist}, \textcolor{red}{red for Pedestrian}, and black for background.
   }
\label{fig:qualitative_results}
\end{figure*}

% ── Section 4.5: original qualitative section — no changes made (minimal-change policy) ──
\subsection{Qualitative Results on CARRADA}

Figure~\ref{fig:qualitative_results} presents qualitative comparisons on two representative scenes from the CARRADA test split, showing the RGB image together with RD and RA semantic predictions from different methods. As seen in both examples, the proposed \textbf{HyperRadar} produces predictions that are more compact, better localised, and more semantically consistent than the competing methods, particularly for sparse foreground objects and weak radar returns.

In the first scene, which contains vehicles with relatively weak and spatially concentrated responses, HyperRadar recovers cleaner foreground structure in both RD and RA views while preserving the extent of the \emph{car} regions more faithfully. Compared with TransRadar, TMVA-Net, MVNet, and U-Net, our predictions exhibit fewer fragmented activations and less background leakage. Competing methods either miss parts of the target, produce overly sparse responses, or introduce noisy foreground regions, whereas HyperRadar remains closer to the ground-truth shape and location across both radar projections. This is especially visible in the RD view, where our method better preserves the main object response while avoiding scattered false positives.

In the second scene, which is more challenging due to the presence of multiple object categories and weaker radar evidence, HyperRadar again yields the most coherent predictions. In particular, it better separates the \emph{pedestrian} and \emph{car} regions and preserves their relative spatial layout in both RD and RA. Baseline methods show stronger confusion between sparse foreground classes, with some methods suppressing small targets and others producing incomplete or noisy masks. By contrast, HyperRadar produces sharper and more stable activations, indicating improved robustness to sparsity and class ambiguity.

Overall, the qualitative results support the quantitative findings: HyperRadar more reliably captures sparse foreground objects, reduces fragmented predictions, and yields masks that are visually closer to the ground truth across both RD and RA views. These visual improvements are consistent with the proposed higher-order structural modelling and cross-view alignment, which help the network maintain coherent object-level predictions under noisy radar sensing.

\section{Limitations and Future Work}

Although \textbf{HyperRadar} achieves strong results, it has some limitations. First, the current formulation operates on 2D RA/RD/AD projections rather than the full RAD tensor, which is efficient but may discard some fine-grained 3D structure. Second, the cross-view regularisation mainly exploits the shared range dimension between RA and RD, providing only a partial geometric constraint and not explicitly modelling longer temporal consistency. Finally, our evaluation is limited to CARRADA and RADIal, which, although widely used, do not cover the full diversity of radar sensors and deployment conditions. Future work will therefore focus on improving efficiency through sparse or hierarchical formulations, extending the method to richer 3D and temporal modelling, and exploring multi-modal and multi-task settings such as radar-camera fusion and joint segmentation, detection, and tracking.

\section{Conclusion}

We introduced \textbf{HyperRadar}, a compact and structurally grounded framework for multi-view radar semantic segmentation. By combining learnable hypergraph-based feature refinement, unbalanced optimal transport for correspondence-free cross-view alignment, and adaptive attention-based fusion, the proposed method addresses key limitations of existing radar segmentation approaches that rely primarily on local or pairwise interactions.

HyperRadar is designed to better match the sparse, noisy, and partially observed nature of automotive radar data, enabling more coherent feature learning within each view and more robust alignment across views. Extensive experiments on CARRADA and RADIal show that this design consistently improves over strong radar-specific baselines, yielding better segmentation accuracy and stronger overall perception performance with an efficient architecture.

These results highlight the value of higher-order relational modelling for radar perception and suggest that structurally aware representation learning is a promising direction for robust scene understanding in adverse sensing conditions. Future work will extend this framework to longer temporal reasoning, multi-modal fusion, and unified perception pipelines for segmentation, detection, and tracking.
\clearpage
\bibliographystyle{splncs04}
\bibliography{main}
\end{document}